\documentclass[11pt,a4paper]{article}
\usepackage[hyperref]{acl2020}
\usepackage{times}
\usepackage{latexsym}

\usepackage{multirow}
\usepackage{graphicx}
\usepackage{subfigure}
\usepackage{enumitem}
\usepackage{color, colortbl}
\usepackage{bm}
\usepackage{amsmath}
\usepackage{amsfonts}
\usepackage{array}
\usepackage{url}
\usepackage{amsthm}
\usepackage{amssymb}

\theoremstyle{definition}

\usepackage{etoolbox}
\newcolumntype{L}[1]{>{\raggedright\arraybackslash}p{#1}}
\newcolumntype{C}[1]{>{\centering\arraybackslash}p{#1}}
\newcolumntype{R}[1]{>{\raggedleft\arraybackslash}p{#1}}
\definecolor{Gray}{gray}{0.9}

\usepackage{algorithm}
\usepackage{algorithmicx}
\usepackage{algpseudocode}
\usepackage{amsmath}

\algnewcommand\algorithmicforeach{\textbf{for each}}
\algdef{S}[FOR]{ForEach}[1]{\algorithmicforeach\ #1\ \algorithmicdo}
\algnewcommand{\LineComment}[1]{\State \(\triangleright\) #1}

\newcommand{\cev}[1]{\reflectbox{\ensuremath{\vec{\reflectbox{\ensuremath{#1}}}}}}


\usepackage{microtype}

\aclfinalcopy 
\def\aclpaperid{1547} 

\title{Semantic Graphs for Generating Deep Questions}

\author{Liangming Pan$^{1,2}$ \quad Yuxi Xie$^{3}$ \quad Yansong Feng$^{3}$\\
\textbf{Tat-Seng Chua$^2$ \quad Min-Yen Kan$^2$}\\
$^1$NUS Graduate School for Integrative Sciences and Engineering\\
$^2$School of Computing, National University of Singapore, Singapore\\
$^3$Wangxuan Institute of Computer Technology, Peking University \\
{\tt e0272310@u.nus.edu, \{xieyuxi, fengyansong\}@pku.edu.cn} \\
{\tt \{dcscts@, kanmy@comp.\}nus.edu.sg}\\
}

\date{}

\begin{document}
\maketitle
\begin{abstract}
This paper proposes the problem of Deep Question Generation (DQG), which aims to generate complex questions that require reasoning over multiple pieces of information of the input passage. 
In order to capture the global structure of the document and facilitate reasoning, we propose a novel framework which first constructs a semantic-level graph for the input document and then encodes the semantic graph by introducing an attention-based GGNN (Att-GGNN). Afterwards, we fuse the document-level and graph-level representations to perform joint training of content selection and question decoding. 
On the HotpotQA deep-question centric dataset, our model greatly improves performance over questions requiring reasoning over multiple facts, leading to state-of-the-art performance. The code is publicly available at
\url{https://github.com/WING-NUS/SG-Deep-Question-Generation}. 
\end{abstract}

\section{Introduction}
\label{sec:Introduction}

\begin{figure}[!t]
	\centering
	\includegraphics[width=7.6cm]{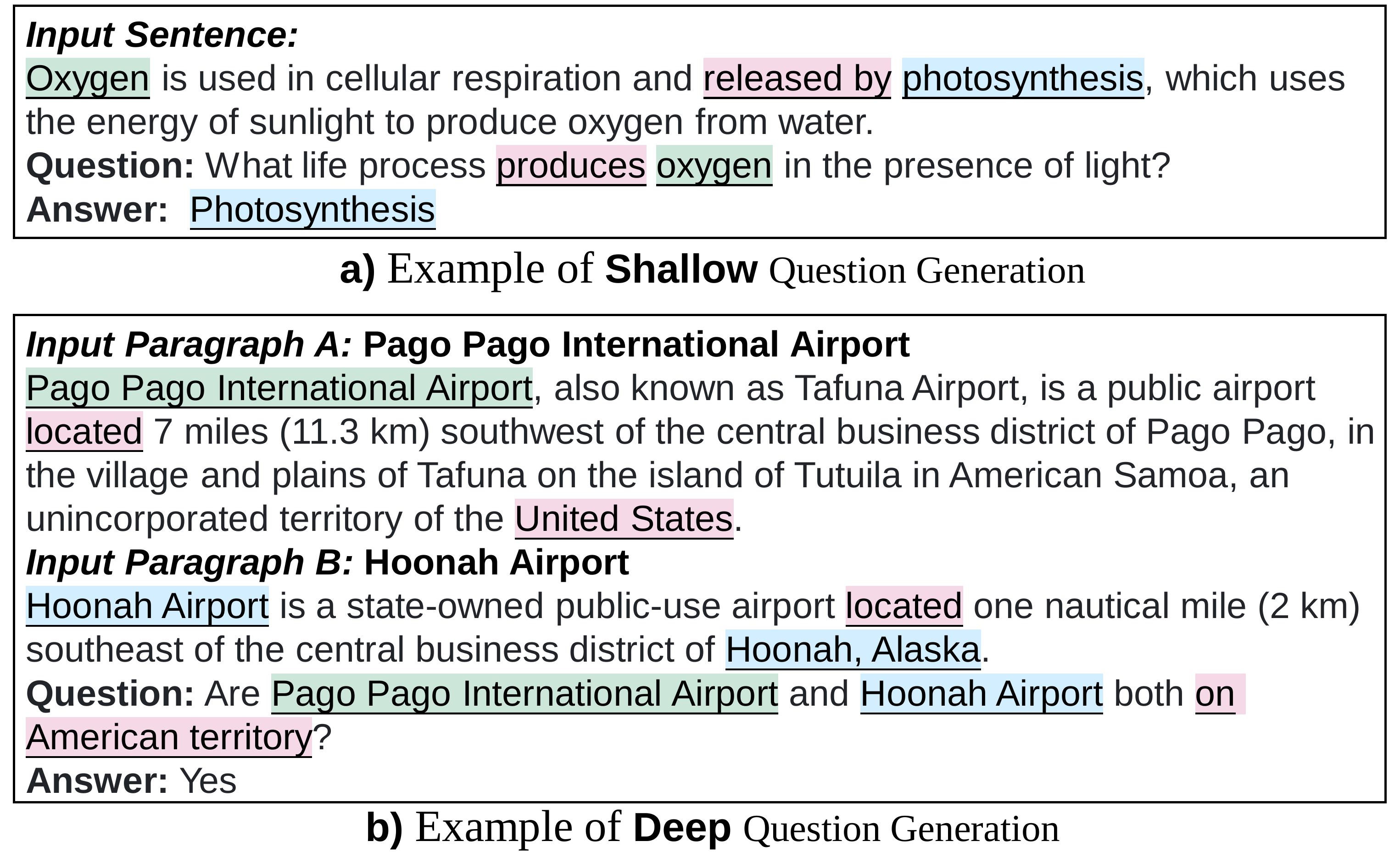}
    \caption{Examples of shallow/deep QG. The evidence needed to generate the question are highlighted.}
    \label{fig:task_example}
\end{figure}

Question Generation (QG) systems play a vital role in question answering (QA), dialogue system, and automated tutoring applications -- by enriching the training QA corpora, helping chatbots start conversations with intriguing questions, and automatically generating assessment questions, respectively. 
Existing QG research has typically focused on generating factoid questions relevant to one fact obtainable from a single sentence~\cite{DBLP:conf/emnlp/DuanTCZ17,DBLP:conf/emnlp/ZhaoNDK18,DBLP:conf/aaai/KimLSJ19}, as exemplified in Figure~\ref{fig:task_example} a). However, less explored has been the comprehension and reasoning aspects of questioning, resulting in questions that are shallow and not reflective of the true creative human process. 

People have the ability to ask deep questions about events, evaluation, opinions, synthesis, or reasons, usually in the form of \textit{Why}, \textit{Why-not}, \textit{How}, \textit{What-if}, which requires an in-depth understanding of the input source and the ability to reason over disjoint relevant contexts; \textit{e.g.}, asking \textit{Why did Gollum betray his master Frodo Baggins?} after reading the fantasy novel \textit{The Lord of the Rings}. Learning to ask such deep questions has intrinsic research value concerning how human intelligence embodies the skills of curiosity and integration, and will have broad application in future intelligent systems. 
Despite a clear push towards {\it answering} deep questions (exemplified by multi-hop reading comprehension~\cite{DBLP:conf/naacl/CaoAT19} and commonsense QA~\cite{DBLP:conf/acl/RajaniMXS19}), {\it generating} deep questions remains un-investigated. There is thus a clear need to push QG research towards generating deep questions that demand higher cognitive skills. 

In this paper, we propose the problem of \textbf{D}eep \textbf{Q}uestion \textbf{G}eneration (DQG), which aims to generate questions that require reasoning over multiple pieces of information in the passage. Figure~\ref{fig:task_example} b) shows an example of deep question which requires a comparative reasoning over two disjoint pieces of evidences. 
DQG introduces three additional challenges that are not captured by traditional QG systems. 
First, unlike generating questions from a single sentence, DQG requires document-level understanding, which may introduce long-range dependencies when the passage is long. 
Second, we must be able to select relevant contexts to ask meaningful questions; this is non-trivial as it involves understanding the relation between disjoint pieces of information in the passage. 
Third, we need to ensure correct reasoning over multiple pieces of information so that the generated question is answerable by information in the passage. 

\begin{figure*}[t]
	\centering
	\includegraphics[width=16cm]{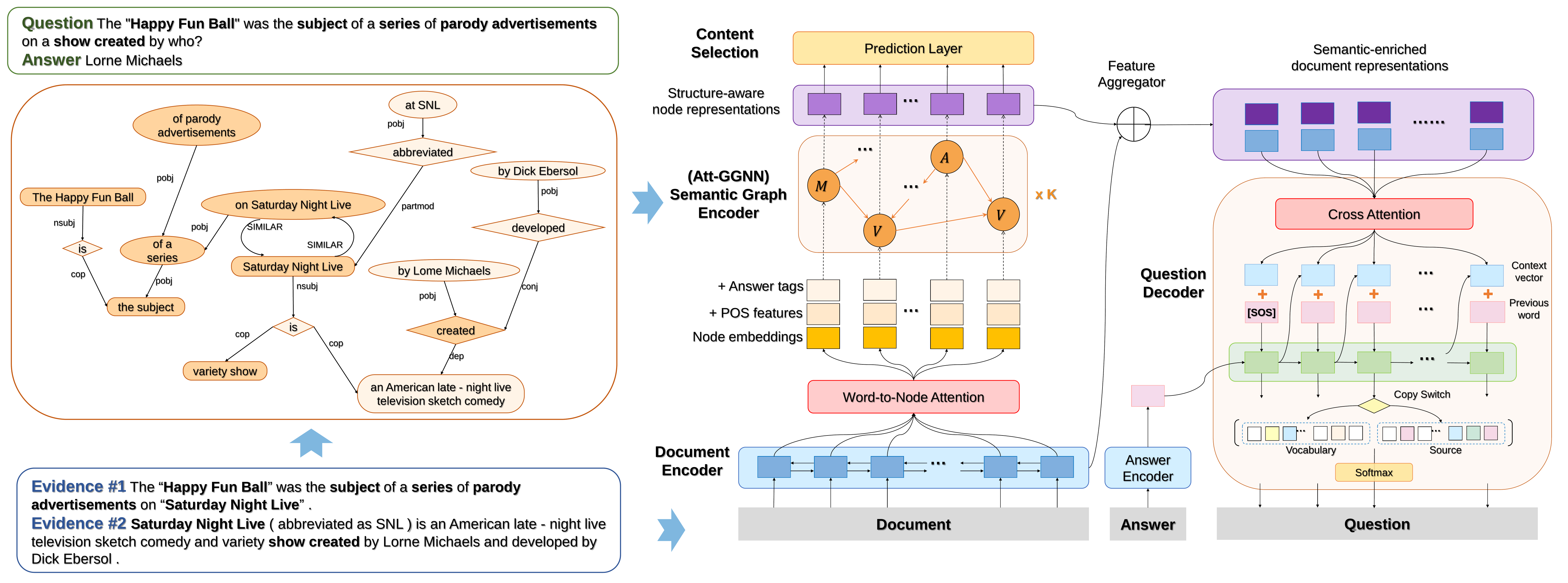}
    \caption{The framework of our proposed model (on the right) together with an input example (on the left). 
    The model consists of four parts: (1) a \textit{document encoder} to encode the input document, (2) a \textit{semantic graph encoder} to embed the document-level semantic graph via Att-GGNN, (3) a \textit{content selector} to select relevant question-worthy contents from the semantic graph, and (4) a \textit{question decoder} to generate question from the semantic-enriched document representation. The left figure shows an input example and its semantic graph. Dark-colored nodes in the semantic graph are question-worthy nodes that are labeled to train the content selection task. 
    }
    \label{fig:model_framework}
\end{figure*}

To facilitate the selection and reasoning over disjoint relevant contexts, we distill important information from the passage and organize them as a \textit{semantic graph}, in which the nodes are extracted based on semantic role labeling or dependency parsing, and connected by different intra- and inter- semantic relations (Figure~\ref{fig:model_framework}). Semantic relations provide important clues about what contents are question-worthy and what reasoning should be performed; \textit{e.g.}, in Figure~\ref{fig:task_example}, both the entities \textit{Pago Pago International Airport} and \textit{Hoonah Airport} have the \textit{located\_at} relation with a city in United States. It is then natural to ask a comparative question: \textit{e.g.}, \textit{Are Pago Pago International Airport and Hoonah Airport both on American territory?}. 
To efficiently leverage the semantic graph for DQG, we introduce three novel mechanisms: 
(1) proposing a novel graph encoder, which incorporates an attention mechanism into the Gated Graph Neural Network (GGNN)~\cite{DBLP:journals/corr/LiTBZ15}, to dynamically model the interactions between different semantic relations; 
(2) enhancing the word-level passage embeddings and the node-level semantic graph representations to obtain an unified semantic-aware passage representations for question decoding; and 
(3) introducing an auxiliary {\it content selection} task that jointly trains with question decoding, which assists the model in selecting relevant contexts in the semantic graph to form a proper reasoning chain. 




We evaluate our model on HotpotQA~\cite{DBLP:conf/emnlp/Yang0ZBCSM18}, a challenging dataset in which the questions are generated by reasoning over text from separate Wikipedia pages. 
Experimental results show that our model --- incorporating both the use of the semantic graph and the content selection task --- improves performance by a large margin, in terms of both automated metrics (Section~\ref{sec:performance_comparison}) and human evaluation (Section~\ref{sec:human_evaluation}). 
Error analysis (Section~\ref{sec:error_analysis}) validates that our use of the semantic graph greatly reduces the amount of semantic errors in generated questions. 
In summary, our contributions are: (1) the very first work, to the best of our knowledge, to investigate deep question generation, (2) a novel framework which combines a semantic graph with the input passage to generate deep questions, and (3) a novel graph encoder that incorporates attention into a GGNN approach. 

\section{Related Work}
\label{sec:Related_Works}

Question generation aims to automatically generate questions from textual inputs.
\textit{Rule-based techniques} for QG usually rely on manually-designed rules or templates to transform a piece of given text to questions~\cite{heilman2011automatic,DBLP:conf/coling/ChaliH12a}. These methods are confined to a variety of transformation rules or templates, making the approach difficult to generalize. \textit{Neural-based approaches} take advantage of the sequence-to-sequence (Seq2Seq) framework with attention~\cite{DBLP:journals/corr/BahdanauCB14}. These models are trained in an end-to-end manner, requiring far less labor and enabling better language flexibility, compared against rule-based methods. A comprehensive survey of QG can be found in~\citet{DBLP:journals/corr/abs-1905-08949}. 

Many improvements have been proposed since the first Seq2Seq model of~\citet{DBLP:conf/acl/DuSC17}: applying various techniques to encode the answer information, thus allowing for better quality answer-focused questions~\cite{DBLP:conf/nlpcc/ZhouYWTBZ17,DBLP:conf/emnlp/SunLLHMW18,DBLP:conf/aaai/KimLSJ19}; improving the training via combining supervised and reinforcement learning to maximize question-specific rewards~\cite{DBLP:conf/rep4nlp/YuanWGSBZST17}; and incorporating various linguistic features into the QG process~\cite{DBLP:conf/www/LiuZNLHWX19}. However, these approaches only consider sentence-level QG.  In contrast, our work focus on the challenge of generating deep questions with multi-hop reasoning over document-level contexts. 

Recently, work has started to leverage paragraph-level contexts to produce better questions. \citet{DBLP:conf/acl/CardieD18} incorporated coreference knowledge to better encode entity connections across documents. 
\citet{DBLP:conf/emnlp/ZhaoNDK18} applied a gated self-attention mechanism to encode contextual information. 
However, in practice, semantic structure is difficult to distil solely via self-attention over the entire document. Moreover, despite considering longer contexts, these works are trained and evaluated on SQuAD~\cite{DBLP:conf/emnlp/RajpurkarZLL16}, which we argue as insufficient to evaluate deep QG because more than 80\% of its questions are shallow and only relevant to information confined to a single sentence~\cite{DBLP:conf/acl/DuSC17}. 

\section{Methodology}
\label{sec:Method}

Given the document $\mathcal{D}$ and the answer $\mathcal{A}$, the objective is to generate a question $\bar{\mathcal{Q}}$ that satisfies: 
\begin{equation}
\label{equ:problem_formulation}
    \bar{\mathcal{Q}} = \arg \max_{\mathcal{Q}} P(\mathcal{Q} \vert \mathcal{D}, \mathcal{A})
\end{equation}
\noindent where document $\mathcal{D}$ and answer $\mathcal{A}$ are both sequences of words. Different from previous works, we aim to generate a $\bar{\mathcal{Q}}$ 
which involves reasoning over multiple evidence sentences $\mathcal{E} = \{s_i\}_{i=1}^n$, where $s_i$ is a sentence in $\mathcal{D}$. Also, unlike traditional settings, $\mathcal{A}$ may not be a sub-span of $\mathcal{D}$ because reasoning is involved to obtain the answer. 

\subsection{General Framework}

We propose an encoder--decoder framework with two novel features specific to DQG: (1) a fused word-level document and node-level semantic graph representation to better utilize and aggregate the semantic information among the relevant disjoint document contexts, and (2) joint training over the question decoding and content selection tasks to improve selection and reasoning over relevant information. 
Figure~\ref{fig:model_framework} shows the general architecture of the proposed model, including three modules: \textit{semantic graph construction}, which builds the DP- or SRL-based semantic graph for the given input; \textit{semantic-enriched document representation}, employing a novel Attention-enhanced Gated Graph Neural Network (Att-GGNN) to learn the semantic graph representations, which are then fused with the input document to obtain graph-enhanced document representations; and \textit{joint-task question generation}, which generates deep questions via joint training of node-level content selection and word-level question decoding. In the following, we describe the details of each module. 

\subsection{Semantic Graph Construction}
\label{sec:semantic_graph_construction}


As illustrated in the introduction, the semantic relations between entities serve as strong clues in determining what to ask about and the reasoning types it includes. To distill such semantic information in the document, we explore both SRL- (Semantic Role Labelling) and DP- (Dependency Parsing) based methods to construct the semantic graph. Refer to Appendix~\ref{sec:appendix} for the details of graph construction. 

\noindent $\bullet$ \textbf{SRL-based Semantic Graph. }
The task of Semantic Role Labeling (SRL) is to identify what semantic relations hold among a predicate and its associated participants and properties~\cite{DBLP:journals/coling/MarquezCLS08}, including ``who" did ``what" to ``whom", etc. 
For each sentence, we extract predicate-argument tuples via SRL toolkits\footnote{We employ the state-of-the-art BERT-based model~\cite{DBLP:journals/corr/abs-1904-05255} in the AllenNLP toolkit to perform SRL. }. Each tuple forms a sub-graph where each tuple element (\textit{e.g.}, arguments, location, and temporal) is a node. We add inter-tuple edges between nodes from different tuples if they have an inclusive relationship or potentially mention the same entity. 



\noindent $\bullet$ \textbf{DP-based Semantic Graph. }
We employ the biaffine attention model~\cite{DBLP:conf/iclr/DozatM17} for each sentence to obtain its dependency parse tree, which is further revised by removing unimportant constituents (\textit{e.g.}, punctuation) and merging consecutive nodes that form a complete semantic unit. Afterwards, we add inter-tree edges between similar nodes from different parse trees to construct a connected semantic graph. 

The left side of Figure~\ref{fig:model_framework} shows an example of the DP-based semantic graph. Compared with SRL-based graphs, DP-based ones typically model more fine-grained and sparse semantic relations, as discussed in Appendix~\ref{appsec:examples}.  Section~\ref{sec:performance_comparison} gives a performance comparison on these two formalisms. 

\subsection{Semantic-Enriched Document Representations}

We separately encode the document $\mathcal{D}$ and the semantic graph $\mathcal{G}$ via an RNN-based passage encoder and a novel Att-GGNN graph encoder, respectively,  then fuse them to obtain the semantic-enriched document representations for question generation. 

\vspace{2mm}

\noindent \textbf{Document Encoding. }
Given the input document $\mathcal{D} = [ w_1, \cdots, w_l ]$, we employ the bi-directional Gated Recurrent Unit (GRU)~\cite{DBLP:conf/emnlp/ChoMGBBSB14} to encode its contexts. We represent the encoder hidden states as $\mathbf{X}_{\mathcal{D}} = [\mathbf{x}_1, \cdots, \mathbf{x}_l]$, where $\mathbf{x}_i = [\vec{\mathbf{x}_i}; \cev{\mathbf{x}_i}]$ is the context embedding of $w_i$ as a concatenation of its bi-directional hidden states. 

\vspace{2mm}

\noindent \textbf{Node Initialization. } We define the SRL- and DP-based semantic graphs in an unified way. 
The semantic graph of the document $\mathcal{D}$ is a heterogeneous multi-relation graph $\mathcal{G} = (\mathcal{V}, \mathcal{E})$, where $\mathcal{V} = \{v_i \}_{i=1:N^v}$ and $\mathcal{E} = \{e_k \}_{k=1:N^e}$ denote graph nodes and the edges connecting them, where $N^v$ and $N^e$ are the numbers of nodes and edges in the graph, respectively. Each node $v = \{w_j\}_{j=m_v}^{n_v}$ is a text span in $\mathcal{D}$ with an associated node type $t_v$, where $m_v$ / $n_v$ is the starting / ending position of the text span. Each edge also has a type $t_e$ that represents the semantic relation between nodes. 

\vspace{2mm}


We obtain the initial representation $\mathbf{h}^0_v$ for each node $v = \{w_j\}_{j=m_v}^{n_v}$ by computing the word-to-node attention. First, we concatenate the last hidden states of the document encoder in both directions as the document representation $\mathbf{d_{\mathcal{D}}}=[\vec{\mathbf{x}}_l; \cev{\mathbf{x}}_1]$. Afterwards, for a node $v$, we calculate the attention distribution of $\mathbf{d_{\mathcal{D}}}$ over all the words $\{ w_{m_v}, \cdots, w_{j}, \cdots, w_{n_v} \}$ in $v$ as follows:

\begin{equation}
\label{equ:word_to_node_attn}
\beta^v_j = \frac{\exp(\text{Attn}( \mathbf{d_{\mathcal{D}}} , \mathbf{x}_j ))}{\sum_{k = m_n}^{n_v} \exp( \text{Attn}(\mathbf{d_{\mathcal{D}}} , \mathbf{x}_k  ) )}
\end{equation}

\noindent where $\beta^v_j$ is the attention coefficient of the document embedding $\mathbf{d_{\mathcal{D}}}$ over a word $w_j$ in the node $v$. The initial node representation $\mathbf{h}^0_v$ is then given by the attention-weighed sum of the embeddings of its constituent words, \textit{i.e.}, $\mathbf{h}^0_v = \sum_{j=m_v}^{n_v} \beta_j^v \mathbf{x}_j$. Word-to-node attention ensures each node to capture not only the meaning of its constituting part but also the semantics of the entire document. The node representation is then enhanced with two additional features: the POS embedding $\mathbf{p}_v$ and the answer tag embedding $\mathbf{a}_v$ to obtain the enhanced initial node representations $\tilde{\mathbf{h}}^0_v = [\mathbf{h}^0_v ; \mathbf{p}_v ; \mathbf{a}_v]$.

\vspace{2mm}

\noindent \textbf{Graph Encoding.} We then employ a novel Att-GGNN to update the node representations by aggregating information from their neighbors. 
To represent multiple relations in the edge, we base our model on the multi-relation Gated Graph Neural Network (GGNN)~\cite{DBLP:journals/corr/LiTBZ15}, which provides a separate transformation matrix for each edge type. For DQG, it is essential for each node to pay attention to different neighboring nodes when performing different types of reasoning. 
To this end, we adopt the idea of Graph Attention Networks~\cite{DBLP:journals/corr/abs-1710-10903} to dynamically determine the weights of neighboring nodes in message passing using an attention mechanism. 

Formally, given the initial hidden states of graph $\mathbf H^0 = \{\tilde{\mathbf{h}}^0_i\} \vert_{v_i \in \mathcal{V}}$, Att-GGNN conducts $K$ layers of state transitions, leading to a sequence of graph hidden states $\mathbf H^0, \mathbf H^1, \cdots, \mathbf H^K$, where $\mathbf H^k = \{\mathbf{h}^{(k)}_j\} \vert_{v_j \in \mathcal{V}}$. 
At each state transition, an aggregation function is applied to each node $v_i$ to collect messages from the nodes directly connected to $v_i$. The neighbors are distinguished by their incoming and outgoing edges as follows:

\begin{equation}
\label{equ:message_passing_out}
\mathbf{h}_{\mathcal{N}_{\vdash(i)}}^{(k)} = \sum_{v_j\in \mathcal{N}_{\vdash(i)}}{\alpha_{ij}^{(k)} \mathbf{W}^{t_{e_{ij}}} \mathbf{h}_{j}^{(k)}}
\end{equation}

\begin{equation}
\label{equ:message_passing_in}
\mathbf{h}_{\mathcal{N}_{\dashv(i)}}^{(k)} = \sum_{v_j\in \mathcal{N}_{\dashv(i)}}{\alpha_{ij}^{(k)} \mathbf{W}^{t_{e_{ji}}} \mathbf{h}_{j}^{(k)}}
\end{equation}

\noindent where $\mathcal{N}_{\dashv(i)}$ and $\mathcal{N}_{\vdash(i)}$ denote the sets of incoming and outgoing edges of $v_i$, respectively. 
$\mathbf{W}^{t_{e_{ij}}}$ denotes the weight matrix corresponding to the edge type $t_{e_{ij}}$ from $v_i$ to $v_j$, and $\alpha_{ij}^{(k)}$ is the attention coefficient of $v_i$ over $v_j$, derived as follows: 

\begin{equation}
\label{equ:GAT}
\alpha^{(k)}_{ij} = \frac{\exp{(\text{Attn}(\mathbf{h}_{i}^{(k)}, \mathbf{h}_{j}^{(k)}))}}{\sum_{t \in \mathcal{N}_{(i)}} \exp(\text{Attn}(\mathbf{h}_{i}^{(k)}, \mathbf{h}_{t}^{(k)}))}
\end{equation}

\noindent where $\text{Attn}(\cdot, \cdot)$ is a single-layer neural network implemented as $\mathbf{a}^T [\mathbf{W}^{A} \mathbf{h}_{i}^{(k)}; \mathbf{W}^{A} \mathbf{h}_{j}^{(k)}]$, here $\mathbf{a}$ and $\mathbf{W}^{A}$ are learnable parameters. 
Finally, an GRU is used to update the node state by incorporating the aggregated neighboring information.

\begin{equation}
\label{equ:GRU_state_update}
\mathbf{h}_{i}^{(k+1)} = \text{GRU} (\mathbf{h}_{i}^{(k)}, \left [ \mathbf{h}_{\mathcal{N}_{\vdash(i)}}^{(k)} \ ; \ \mathbf{h}_{\mathcal{N}_{\dashv(i)}}^{(k)} \right ])
\end{equation}

\noindent After the $K$-th state transition, we denote the final structure-aware representation of node $v$ as $\mathbf{h}^K_v$. 

\vspace{2mm}

\noindent \textbf{Feature Aggregation.} Finally, we fuse the semantic graph representations $\mathbf{H}^K$ with the document representations $\mathbf{X}_{\mathcal{D}}$ to obtain the semantic-enriched document representations $\mathbf{E}_{\mathcal{D}}$ for question decoding, as follows:

\begin{equation}
\label{equ:feature_fusion}
\mathbf{E}_{\mathcal{D}} = \text{Fuse}(\mathbf{X}_{\mathcal{D}}, \mathbf{H}^K)
\end{equation}

We employ a simple matching-based strategy for the feature fusion function $\text{Fuse}$. For a word $w_i \in \mathcal{D}$, we match it to the smallest granularity node that contains the word $w_i$, denoted as $v_{M(i)}$. We then concatenate the word representation $\mathbf{x}_i$ with the node representation $\mathbf{h}^K_{v_{M(i)}}$, \textit{i.e.}, $\mathbf{e}_i = [\mathbf{x}_i \ ; \ \mathbf{h}^K_{v_{M(i)}}]$. When there is no corresponding node $v_{M(i)}$, we concatenate $\mathbf{x}_i$ with a special vector close to $\vec{0}$.

The semantic-enriched representation $\mathbf{E}_{\mathcal{D}}$ provides the following important information to benefit question generation: (1) \textit{semantic information}: the document incorporates semantic information explicitly through concatenating with semantic graph encoding; 
(2) \textit{phrase information}: a phrase is often represented as a single node in the semantic graph (\textit{cf} Figure~\ref{fig:model_framework} as an example); therefore its constituting words are aligned with the same node representation; 
(3) \textit{keyword information}: a word (\textit{e.g.}, a preposition) not appearing in the semantic graph is aligned with the special node vector mentioned before, indicating the word does not carry important information. 

\subsection{Joint Task Question Generation}
\label{sec:joint_task}

Based on the semantic-rich input representations, we generate questions via jointly training on two tasks: \textit{Question Decoding} and \textit{Content Selection}. 

\vspace{2mm}

\noindent \textbf{Question Decoding. }We adopt an attention-based GRU model~\cite{DBLP:journals/corr/BahdanauCB14} with copying~\cite{DBLP:conf/acl/GuLLL16,DBLP:conf/acl/SeeLM17} and coverage mechanisms~\cite{DBLP:conf/acl/TuLLLL16} as the question decoder. 
The decoder takes the semantic-enriched representations $\mathbf{E}_{\mathcal{D}} = \{ \mathbf{e}_i, \forall w_i \in \mathcal{D} \}$ from the encoders as the attention memory to generate the output sequence one word at a time. 
To make the decoder aware of the answer, we use the average word embeddings in the answer to initialize the decoder hidden states. 

At each decoding step $t$, the model learns to attend over the input representations $\mathbf{E}_{\mathcal{D}}$ and compute a context vector $\mathbf{c}_t$ based on $\mathbf{E}_{\mathcal{D}}$ and the current decoding state $\mathbf{s}_t$. 
Next, the copying probability $P_{cpy} \in [0,1]$ is calculated from the context vector $\mathbf{c}_t$, the decoder state $\mathbf{s}_t$ and the decoder input $y_{t-1}$. $P_{cpy}$ is used as a soft switch to choose between generating from the vocabulary, or copying from the input document.
Finally, we incorporate the coverage mechanisms~\cite{DBLP:conf/acl/TuLLLL16} to encourage the decoder to utilize diverse components of the input document. Specifically, at each step, we maintain a coverage vector $\mathbf{cov}_t$, which is the sum of attention distributions over all previous decoder steps. A coverage loss is computed to penalize repeatedly attending to the same locations of the input document. 

\vspace{2mm}
\noindent \textbf{Content Selection. }
To raise a deep question, humans select and reason over relevant content. 
To mimic this, we propose an auxiliary task of content selection to jointly train with question decoding. We formulate this as a node classification task, \textit{i.e.}, deciding whether each node should be involved in the process of asking, \textit{i.e.}, appearing in the reasoning chain for raising a deep question, exemplified by the dark-colored nodes in Figure~\ref{fig:model_framework}. 

To this end, we add one feed-forward layer on top of the final-layer of the graph encoder, taking the output node representations $\mathbf{H}^K$ for classification. We deem a node as positive ground-truth to train the content selection task if its contents appear in the ground-truth question or act as a bridge entity between two sentences. 

Content selection helps the model to identify the question-worthy parts that form a proper reasoning chain in the semantic graph. This synergizes with the question decoding task which focuses on the fluency of the generated question. We jointly train these two tasks with weight sharing on the input representations.  

\section{Experiments}
\label{sec:Experiments}

\subsection{Data and Metrics}
To evaluate the model's ability to generate deep questions, we conduct experiments on HotpotQA~\cite{DBLP:conf/emnlp/Yang0ZBCSM18}, containing $\sim$100K crowd-sourced questions that require reasoning over separate Wikipedia articles. Each question is paired with two supporting documents that contain the evidence necessary to infer the answer. In the DQG task, we take the supporting documents along with the answer as inputs to generate the question. 
However, state-of-the-art semantic parsing models have difficulty in producing accurate semantic graphs for very long documents. We therefore pre-process the original dataset to select relevant sentences, \textit{i.e.}, the evidence statements and the sentences that overlap with the ground-truth question, as the input document. 
We follow the original data split of HotpotQA to pre-process the data, resulting in 90,440 / 6,072 examples for training and evaluation, respectively. 



\begin{table*}[!t]
    \small
	\begin{center}
		\begin{tabular}{ c | l | c | c | c | c | c | c } \hline
        \multicolumn{2}{c|}{\textbf{Model}} & \textbf{BLEU1} & \textbf{BLEU2} & \textbf{BLEU3} & \textbf{BLEU4} & \textbf{METEOR} & \textbf{ROUGE-L} \\ \hline \hline
        \multirow{6}{*}{Baselines} 
        & B1. Seq2Seq + Attn & 32.97 & 21.11 & 15.41 & 11.81 & 18.19 & 33.48 \\
        & B2. NQG++ & 35.31 & 22.12 & 15.53 & 11.50 & 16.96 & 32.01 \\
        & B3. ASs2s & 34.60 & 22.77 & 15.21 & 11.29 & 16.78 & 32.88 \\ 
        & B4. S2s-at-mp-gsa & 35.36 & 22.38 & 15.88 & 11.85 & 17.63 & 33.02 \\
        & B5. S2s-at-mp-gsa (+cov, +ans) & 38.74 & 24.89 & 17.88 & 13.48 & 18.39 & 34.51 \\
        & B6. CGC-QG & 31.18 & 22.55 & 17.69 & 14.36 & \textbf{25.20} & \textbf{40.94} \\
        \hline
        \multirow{2}{*}{Proposed} 
        & P1. SRL-Graph & 40.40 & 26.83 & 19.66 & 15.03 & 19.73 & 36.24 \\
        & P2. DP-Graph & \textbf{40.55} & \textbf{27.21} & \textbf{20.13} & \textbf{15.53} & 20.15 & 36.94 \\ \hline \hline
        \multirow{5}{*}{Ablation} 
        & A1. \textit{-w/o} Contexts & 36.48 & 20.56 & 12.89 & 8.46 & 15.43 & 30.86 \\
        & A2. \textit{-w/o} Semantic Graph & 37.63 & 24.81 & 18.14 & 13.85 & 19.24 & 34.93 \\
        & A3. \textit{-w/o} Multi-Relation \& Attention & 38.50 & 25.37 & 18.54 & 14.15 & 19.15 & 35.12 \\
        & A4. \textit{-w/o} Multi-Task & 39.43 & 26.10 & 19.14 & 14.66 & 19.25 & 35.76 \\ \hline
		\end{tabular}
	\end{center}
\caption{Performance comparison with baselines and the ablation study. The best performance is in bold.}
\label{tbl:performance_comparision}
\end{table*}

Following previous works, we employ BLEU 1--4~\cite{DBLP:conf/acl/PapineniRWZ02}, METEOR~\cite{DBLP:conf/wmt/LavieA07}, and ROUGE-L~\cite{lin2004rouge} as automated evaluation metrics. BLEU measures the average $n$-gram overlap on a set of reference sentences.  Both METEOR and ROUGE-L specialize BLEU's n-gram overlap idea for machine translation and text summarization evaluation, respectively.  Critically, we also conduct human evaluation, where annotators evaluate the generation quality from three important aspects of deep questions: fluency, relevance, and complexity. 

\subsection{Baselines}
We compare our proposed model against several strong baselines on question generation. 

\vspace{2mm}
\noindent $\bullet$ \textbf{Seq2Seq + Attn}~\cite{DBLP:journals/corr/BahdanauCB14}: the basic Seq2Seq model with attention, which takes the document as input to decode the question. 

\noindent $\bullet$ \textbf{NQG++}~\cite{DBLP:conf/nlpcc/ZhouYWTBZ17}: which enhances the Seq2Seq model with a feature-rich encoder containing answer position, POS and NER information. 

\noindent $\bullet$ \textbf{ASs2s}~\cite{DBLP:conf/aaai/KimLSJ19}: learns to decode questions from an answer-separated passage encoder together with a keyword-net based answer encoder. 

\noindent $\bullet$ \textbf{S2sa-at-mp-gsa}~\cite{DBLP:conf/emnlp/ZhaoNDK18}: an enhanced Seq2Seq model incorporating gated self-attention and maxout-pointers to encode richer passage-level contexts (B4 in Table~\ref{tbl:performance_comparision}). We also implement a version that uses coverage mechanism and our answer encoder for fair comparison, labeled B5.


\noindent $\bullet$ \textbf{CGC-QG}~\cite{DBLP:conf/www/LiuZNLHWX19}: another enhanced Seq2Seq model that performs word-level content selection before generation; \textit{i.e.}, making decisions on which words to generate and to copy using rich syntactic features, such as NER, POS, and DEP. 



\vspace{2mm}

\noindent {\bf Implementation Details.} For fair comparison, we use the original implementations of ASs2s and CGC-QG to apply them on HotpotQA. All baselines share a 1-layer GRU document encoder and question decoder with hidden units of 512 dimensions. Word embeddings are initialized with 300-dimensional pre-trained GloVe~\cite{DBLP:conf/emnlp/PenningtonSM14}. For the graph encoder, the node embedding size is 256, plus the POS and answer tag 
embeddings with $32$-D for each. The number of layers $K$ is set to 3 and hidden state size is 256. Other settings for training follow standard best practice\footnote{All models are trained using Adam~\cite{DBLP:journals/corr/KingmaB14} with mini-batch size $32$.  
The learning rate is initially set to $0.001$, and adaptive learning rate decay applied. We adopt early stopping and the dropout rate is set to $0.3$ for both encoder and decoder and $0.1$ for all attention mechanisms.}. 


\begin{table*}[!t]
    \small
	\begin{center}
		\begin{tabular}{ l | c | c | c || c | c | c || c | c | c || c | c | c } \hline
		\multirow{2}{*}{Model} & \multicolumn{3}{c||}{Short Contexts} & \multicolumn{3}{c||}{Medium Contexts} & \multicolumn{3}{c||}{Long Contexts} & \multicolumn{3}{c}{Average}\\ \cline{2-13}
        & Flu. & Rel. & Cpx. & Flu. & Rel. & Cpx. & Flu. & Rel. & Cpx. & Flu. & Rel. & Cpx. \\ \hline \hline
        B4. S2sa-at-mp-gsa & 3.76 & 4.25 & 3.98 & 3.43 & 4.35 & 4.13 & 3.17 & 3.86 & 3.57 & 3.45 & 4.15 & 3.89 \\
        B6. CGC-QG & 3.91 & 4.43 & 3.60 & 3.63 & 4.17 & 4.10 & \textbf{3.69} & 3.85 & \textbf{4.13} & 3.75 & 4.15 & 3.94 \\
        A2. \textit{-w/o} Semantic Graph & 4.01 & 4.43 & 4.15 & 3.65 & 4.41 & 4.12 & 3.54 & 3.88 & 3.55 & 3.73 & 4.24 & 3.94 \\
        A4. \textit{-w/o} Multi-Task & 4.11 & 4.58 & 4.28 & 3.81 & 4.27 & \textbf{4.38} & 3.44 & 3.91 & 3.84 & 3.79 & 4.25 & 4.17 \\
        P2. DP-Graph & \textbf{4.34} & \textbf{4.64} & \textbf{4.33} & \textbf{3.83} & \textbf{4.51} & 4.28 & 3.55 & \textbf{4.08} & 4.04 & \textbf{3.91} & \textbf{4.41} & \textbf{4.22} \\ \hline \hline
        G1. Ground Truth & 4.75 & 4.87 & 4.74 & 4.65 & 4.73 & 4.73 & 4.46 & 4.61 & 4.55 & 4.62 & 4.74 & 4.67 \\ \hline
		\end{tabular}
	\end{center}
\vspace{-2mm}
\caption{Human evaluation results for different methods on inputs with different lengths. \textit{Flu.}, \textit{Rel.}, and \textit{Cpx.} denote the \textit{Fluency}, \textit{Relevance}, and \textit{Complexity}, respectively. Each metric is rated on a 1--5 scale (5 for the best). }
\label{tbl:human_evaluation}
\vspace{-2mm}
\end{table*}

\subsection{Comparison with Baseline Models}
\label{sec:performance_comparison}

The top two parts of Table~\ref{tbl:performance_comparision} show the experimental results comparing against all baseline methods. We make three main observations:

1. The two versions of our model --- P1 and P2 --- consistently outperform all other baselines in BLEU. Specifically, our model with DP-based semantic graph (P2) achieves an absolute improvement of 2.05 in BLEU-4 ($+15.2\%$), compared to the document-level QG model which employs gated self-attention and has been enhanced with the same decoder as ours (B5). This shows the significant effect of semantic-enriched document representations, equipped with auxiliary content selection for generating deep questions. 

2. The results of CGC-QG (B6) exhibits an unusual pattern compared with other methods, achieving the \textit{best} METEOR and ROUGE-L but \textit{worst} BLEU-1 among all baselines. As CGC-QG performs word-level content selection, we observe that it tends to include many irrelevant words in the question, leading to lengthy questions ($33.7$ tokens on average, while $17.7$ for ground-truth questions and $19.3$ for our model) that are unanswerable or with semantic errors. Our model greatly reduces the error with node-level content selection based on semantic relations (shown in Table~\ref{tbl:error_analysis}). 


3. While both SRL-based and DP-based semantic graph models (P1 and P2) achieve state-of-the-art BLEU, DP-based graph (P2) performs slightly better ($+3.3\%$ in BLEU-4). A possible explanation is that SRL fails to include fine-grained semantic information into the graph, as the parsing often results in nodes containing a long sequence of tokens. 

\subsection{Ablation Study}

We also perform ablation studies to assess the impact of different components 
on the model performance against our DP-based semantic graph (P2) model.  These are shown as Rows A1--4 in Table~\ref{tbl:performance_comparision}. Similar results are observed for the SRL-version. 

\vspace{2mm}
\noindent $\bullet$ \textbf{Impact of semantic graph. }
When we do not employ the semantic graph (A2, \textit{-w/o} Semantic Graph), the BLEU-4 score of our model dramatically drops to $13.85$, which indicates the necessity of building semantic graphs to model semantic relations between relevant content for deep QG. 
Despite its vital role, result of A1 shows that generating questions purely from the semantic graph is unsatisfactory. We posit three reasons: 1) the semantic graph alone is insufficient to convey the meaning of the entire document, 
2) sequential information in the passage is not captured by the graph, and that 3) the automatically built semantic graph inevitably contains much noise.  These reasons necessitate the composite document representation. 

\vspace{2mm}
\noindent $\bullet$ \textbf{Impact of Att-GGNN.} Using a normal GGNN (A3, \textit{-w/o} Multi-Relation \& Attention) to encode the semantic graph, performance drops to 14.15 ($-3.61\%$) in BLEU-4 compared to the model with Att-GGNN (A4, \textit{-w/o} Multi-Task). 
This reveals that different entity types and their semantic relations provide
auxiliary 
information needed to generate meaningful questions. 
Our Att-GGNN model (P2) incorporates attention into the normal GGNN, effectively leverages the information across multiple node and edge types. 

\vspace{2mm}
\noindent $\bullet$ \textbf{Impact of joint training. }By turning off the content selection task (A4, \textit{-w/o} Multi-Task), the BLEU-4 score drops from $15.53$ to $14.66$, showing the contribution of joint training with the auxiliary task of content selection. We further show that content selection helps to learn a QG-aware graph representation in Section~\ref{sec:case_study}, which trains the model to focus on the question-worthy content and form a correct reasoning chain in question decoding. 

\begin{table*}
\scriptsize
	\begin{center}
		\begin{tabular}{ c | c | l | c | c | c } \hline
			\multirow{2}{*}{Types} & \multicolumn{2}{c|}{\multirow{2}{*}{Examples}} & S2sa-at- & \multirow{2}{*}{CGC-QG} & \multirow{2}{*}{DP-Graph} \\
			& \multicolumn{2}{c|}{$ $} & mp-gsa & & \\ \hline \hline
			\multirow{2}{*}{Correct} & (Pred.) & Between Kemess Mine and Colomac Mine, which mine was operated earlier? & \multirow{2}{*}{56.5\%} & \multirow{2}{*}{52.9\%} &  \multirow{2}{*}{67.4\%} \\
			& (G.T.) & What mine was operated at an earlier date, Kemess Mine or Colomac Mine? & & & \\\hline
			Semantic & (Pred.) & Lawrence Ferlinghetti is an American poet, \textbf{he is} a short story \textbf{written by who}? & \multirow{2}{*}{17.7\%} & \multirow{2}{*}{26.4\%} & \multirow{2}{*}{8.3\%} \\
			Error & (G.T.) & Lawrence Ferlinghetti is an American poet, he wrote a short story named what ? & & & \\ \hline
			Answer & (Pred.) & What is the release date of this game \textbf{released on 17 October 2006}? & \multirow{2}{*}{2.1\%} & \multirow{2}{*}{5.7\%} & \multirow{2}{*}{1.4\%} \\
			Revealing & (G.T.) & What is the release date of this game named Hurricane? & & & \\\hline
            Ghost & (Pred.) & When was the video game on which \textbf{Michael Gelling} plays Dr. Promoter? & \multirow{2}{*}{6.8\%} & \multirow{2}{*}{0.7\%} &  \multirow{2}{*}{4.9\%} \\
			Entity & (G.T.) & When was the video game on which Drew Gelling plays Dr. Promoter? & & & \\\hline
			\multirow{2}{*}{Redundant} & (Pred.) & What town did \textbf{Walcha and Walcha} belong to? & \multirow{2}{*}{16.3\%} & \multirow{2}{*}{14.3\%} & \multirow{2}{*}{13.9\%} \\
			 & (G.T.) & What town did Walcha belong to? & & &
			 \\\hline
			\multirow{2}{*}{Unanswerable} & (Pred.) & What is \textbf{the population of} the city Barack Obama was born? & \multirow{2}{*}{8.2\%} & \multirow{2}{*}{18.6\%} & \multirow{2}{*}{8.3\%} \\
			& (G.T.) & What was the ranking of the population of the city Barack Obama was born in 1999? & & & \\\hline
		\end{tabular}
	\end{center}
\caption{Error analysis on $3$ different methods, with respects to $5$ major error types (excluding the ``Correct"). \textbf{Pred.} and \textbf{G.T.} show the example of the predicted question and the ground-truth question, respectively. \textbf{Semantic Error}: the question has logic or commonsense error; \textbf{Answer Revealing}: the question reveals the answer; \textbf{Ghost Entity}: the question refers to entities that do not occur in the document; \textbf{Redundant}: the question contains unnecessary repetition; \textbf{Unanswerable}: the question does not have the above errors but cannot be answered by the document. }
\label{tbl:error_analysis}
\end{table*}

\subsection{Human Evaluation}
\label{sec:human_evaluation}

We conduct human evaluation on 300 random test samples consisting of: 100 short ($<$50 tokens), 100 medium (50-200 tokens), and 100 long ($>$200 tokens) documents. 
We ask three workers to rate the 300 generated questions as well as the ground-truth questions between 1 (poor) and 5 (good) on three criteria: (1) \textit{Fluency}, which indicates whether the question follows the grammar and accords with the correct logic; (2) \textit{Relevance}, which indicates whether the question is answerable and relevant to the passage; (3) \textit{Complexity}, which indicates whether the question involves reasoning over multiple sentences from the document. We average the scores from raters on each question and report the performance over five top models from Table~\ref{tbl:performance_comparision}. Raters were unaware of the identity of the models in advance. 
Table~\ref{tbl:human_evaluation} shows our human evaluation results, which further validate that our model generates questions of better quality than the baselines. Let us explain two observations in detail: 

\noindent $\bullet$ Compared against B4 (S2sa-at-mp-gsa), improvements are more salient in terms of ``Fluency" ($+13.33\%$) and ``Complexity" ($+8.48\%$) than that of ``Relevance" ($+6.27\%$). The reason is that the baseline produces more shallow questions (affecting complexity) or questions with semantic errors (affecting fluency). We observe similar results when removing the semantic graph (A2. \textit{-w/o} Semantic Graph). These demonstrate that our model, by incorporating the semantic graph, produces questions with fewer semantic errors and utilizes more context. 

\noindent $\bullet$ All metrics decrease in general when the input document becomes longer, with the most obvious drop in ``Fluency". When input contexts is long, it becomes difficult for models to capture question-worthy points and conduct correct reasoning, leading to more semantic errors. Our model tries to alleviate this problem by introducing semantic graph and content selection, but question quality drops as noise increases in the semantic graph when the document becomes longer. 


\subsection{Error Analysis}
\label{sec:error_analysis}
\begin{figure*}[!t]
	\centering
	\includegraphics[width=16cm]{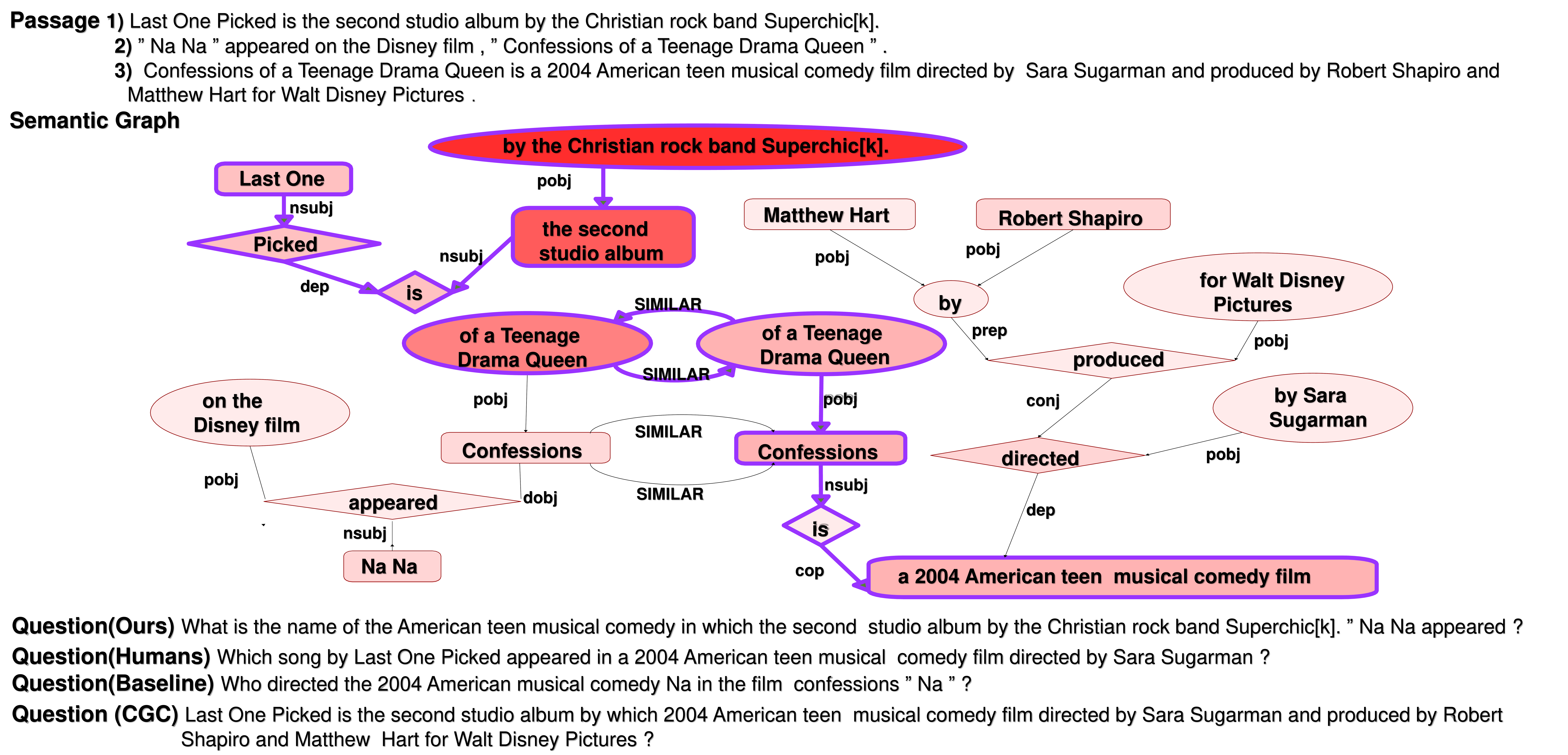}
    \caption{An example of generated questions and average attention distribution on the semantic graph, with nodes colored darker for more attention (best viewed in color). }
    \label{fig:case_study_example}
\end{figure*}

In order to better understand the question generation quality, we manually check the sampled outputs, and list the $5$ main error sources in Table~\ref{tbl:error_analysis}. Among them, ``Semantic Error", ``Redundant", and ``Unanswerable" are noticeable errors for all models. However, we find that baselines have more unreasonable subject--predicate--object collocations (semantic errors) than our model. Especially, CGC-QG (B6) has the largest semantic error rate of $26.4\%$ among the three methods; it tends to copy irrelevant contents from the input document. Our model greatly reduces such semantic errors to $8.3\%$, as we explicitly model the semantic relations between entities by introducing typed semantic graphs. The other noticeable error type is ``Unanswerable"; \textit{i.e.}, the question is correct itself but cannot be answered by the passage. Again, CGC-QG remarkably produces more unanswerable questions than the other two models, and our model achieves comparable results with S2sa-at-mp-gsa (B4), likely due to the fact that answerability requires a deeper understanding of the document as well as commonsense knowledge. These issues cannot be fully addressed by incorporating semantic relations. Examples of questions generated by different models are shown in Figure~\ref{fig:case_study_example}. 

\subsection{Analysis of Content Selection}
\label{sec:case_study}

We introduced the content selection task to guide the model to \textit{select relevant content} and \textit{form proper reasoning chains} in the semantic graph. 
To quantitatively validate the relevant content selection, we calculate the alignment of node attention $\alpha_{v_i}$ with respect to the relevant nodes $\sum_{v_i \in RN}\alpha_{v_i}$ and irrelevant nodes $\sum_{v_i \notin RN}\alpha_{v_i}$, respectively, under the conditions of both single training and joint training, where $RN$ represents the ground-truth we set for content selection. 
Ideally, a successful model should focus on relevant nodes and ignore irrelevant ones; this is reflected by the ratio between $\sum_{v_i \in RN}\alpha_{v_i}$ and $\sum_{v_i \notin RN}\alpha_{v_i}$. 

When jointly training with content selection, this ratio is $1.214$ compared with $1.067$ under single-task training, consistent with our intuition about content selection. Ideally, a successful model should concentrate on parts of the graph that help to form proper reasoning. 
To quantitatively validate this, we compare the concentration of attention in single- and multi-task settings by computing the entropy $H=-\sum \alpha_{v_i}\log{\alpha_{v_i}}$ of the attention distributions. 
We find that content selection increases the entropy from 3.51 to 3.57 on average. To gain better insight, in Figure~\ref{fig:case_study_example}, we visualize the semantic graph attention distribution of an example. 
We see that the model pays more attention (is darker) to the nodes that form the reasoning chain (the highlighted paths in purple), consistent with the quantitative analysis.

\section{Conclusion and Future Works}
\label{sec:FutureDirections}

We propose the problem of DQG to generate questions that requires reasoning over multiple disjoint pieces of information. 
To this end, we propose a novel framework which incorporates semantic graphs to enhance the input document representations and generate questions by jointly training with the task of content selection. Experiments on the HotpotQA dataset demonstrate that introducing semantic graph significantly reduces the semantic errors, and content selection benefits the selection and reasoning over disjoint relevant contents, leading to questions with better quality. 

There are at least two potential future directions. First, graph structure that can accurately represent the semantic meaning of the document is crucial for our model. Although DP-based and SRL-based semantic parsing are widely used, more advanced semantic representations could also be explored, such as discourse structure representation~\cite{DBLP:journals/tacl/NoordATB18,DBLP:conf/acl/LiuCL19} and knowledge graph-enhanced text representations~\cite{DBLP:conf/acl/CaoHJCL17,DBLP:conf/cvpr/YangTZC19}. Second, our method can be improved by explicitly modeling the reasoning chains in generation of deep questions, inspired by related methods~\cite{DBLP:conf/emnlp/LinSX18,DBLP:conf/emnlp/JiangB19} in multi-hop question answering. 


\section*{Acknowledgments}

This research is supported by the National Research Foundation, Singapore under its International Research Centres in Singapore Funding Initiative. Any opinions, findings and conclusions or recommendations expressed in this material are those of the author(s) and do not reflect the views of National Research Foundation, Singapore. 

\bibliography{acl2020}
\bibliographystyle{acl_natbib}

\appendix

\clearpage

\section{Supplemental Material}
\label{sec:appendix}

Here we give a more detailed description for the semantic graph construction, where we have employed two methods: \textit{Semantic Role Labelling (SRL)} and \textit{Dependency Parsing (DP)}. 

\subsection{SRL-based Semantic Graph}
\label{appsec:SRL}
The primary task of semantic role labeling (SRL) is to indicate exactly what semantic relations hold among a predicate and its associated participants and properties~\cite{Marquez:2008:SRL:1403157.1403158}.
Given a document $\mathcal{D}$ with $n$ sentences $\{s_1, \cdots, s_n\}$, Algorithm 1 gives the detailed procedure of constructing the semantic graph based on SRL. 

\begin{algorithm}
    \caption{Build SRL-based Semantic Graphs}
    \begin{algorithmic}[1]
    \label{alg:SRL_graph}
    \Require{Document $\mathcal{D} = \{s_1,\cdots, s_n\}$}
    \Ensure{Semantic graph $\mathcal{G}$}
        \State \Comment{build SRL graph} 
        \State $\mathcal{D} \gets \Call{coreference\_resolution}{\mathcal{D}}$
        \State $\mathcal{G} = \{ \mathcal{V}, \mathcal{E} \}, \mathcal{V} \gets \emptyset, \mathcal{E} \gets \emptyset$
        \ForEach {sentence $s$ \textbf{in} $\mathcal{D}$}
            \State $\mathcal{S} \gets \Call{semantic\_role\_labeling}{s}$
            \ForEach {tuple $\mathbf{t} = (a, v, m)$ \textbf{in} $\mathcal{S}$}
                \State $\mathcal{V}, \mathcal{E} \gets \Call{update\_links}{\mathbf{t}, \mathcal{V}, \mathcal{E}}$
                \State $\mathcal{V} \gets \mathcal{V} \cup \{ a, v, m \}$
                \State $\mathcal{E} \gets \mathcal{E} \cup \{ \langle a, r^{a\rightarrow v}, v \rangle, \langle v, r^{v\rightarrow m}, m \rangle \}$
            \EndFor
        \EndFor
        
        \State \Comment{link to existing nodes}
        \Procedure{update\_links}{$\mathbf{t}, \mathcal{V}, \mathcal{E}$}
            \ForEach {element $e$ \textbf{in} $\mathbf{t}$}
                \ForEach {node $v_i$ \textbf{in} $\mathcal{V}$}
                    \If{\Call{Is\_Similar}{$v_i$, $e$}}
                        \State $\mathcal{E} \gets \mathcal{E} \cup \{ \langle e, r^s, v_i \rangle \}$
                        \State $\mathcal{E} \gets \mathcal{E} \cup \{ \langle v_i, r^s, e \rangle \}$
                    \EndIf
                \EndFor
            \EndFor
        \EndProcedure \\
        \Return {$\mathcal{G}$}
    \end{algorithmic}
\end{algorithm}

We first create an empty graph $\mathcal{G} = (\mathcal{V}, \mathcal{E})$, where $\mathcal{V}$ and $\mathcal{E}$ are the node and edge sets, respectively. For each sentence $s$, we use the state-of-the-art BERT-based model~\cite{DBLP:journals/corr/abs-1904-05255} provided in the AllenNLP toolkit\footnote{https://demo.allennlp.org/semantic-role-labeling} to perform SRL, resulting a set of SRL tuples $\mathcal{S}$. Each tuple $\mathbf{t} \in \mathcal{S}$ consists of an argument $a$, a verb $v$, and (possibly) a modifier $m$, each of which is a text span of the sentence. We treat each of $a$, $v$, and $m$ as a node and link it to an existing node $v_i \in \mathcal{V}$ if it is similar to $v_i$. Two nodes $A$ and $B$ are similar if one of following rules are satisfied: (1) $A$ is equal to $B$; (2) $A$ contains $B$; (3) the number of overlapped words between $A$ and $B$ is larger than the half of the minimum number of words in $A$ and $B$. The edge between two similar nodes is associated with a special semantic relationship \textit{SIMILAR}, denoted as $r^s$. Afterwards, we add two edges $\langle a, r^{a\rightarrow v}, v \rangle$ and $\langle v, r^{v\rightarrow m}, m \rangle$ into the edge set, where $r^{a\rightarrow v}$ and $r^{v\rightarrow m}$ denotes the semantic relationship between $(a, v)$ and $(v, w)$, respectively. As a result, we obtain a semantic graph with multiple node and edge types based on the SRL, which captures the core semantic relations between entities within the document. 

\begin{figure*}[!t]
	\centering
	\includegraphics[width=16cm]{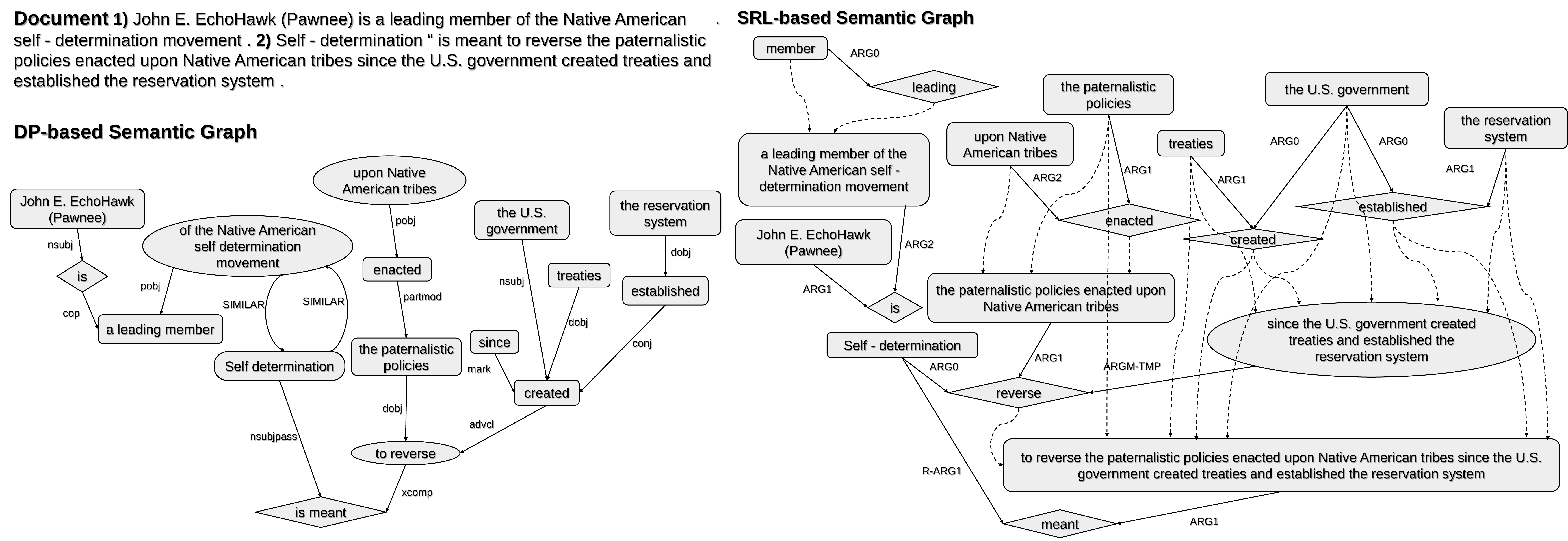}
    \caption{An example of constructed DP- and SRL- based semantic graphs, where $\dashrightarrow$ indicates \textit{CHILD} relation, and rectangular, rhombic and circular nodes represent arguments, verbs and modifiers respectively.}
    \label{fig:appenix1}
\end{figure*}

\begin{algorithm}
    \caption{Build DP-based Semantic Graphs}
    \begin{algorithmic}[1]
    \Require{Document $\mathcal{D} = \{s_1,\cdots, s_n\}$}
    \Ensure{Semantic graph $\mathcal{G}$}
        \State \Comment{Dependency parsing} 
        \State $\mathcal{T} \gets \emptyset$ 
        \State $\mathcal{D} \gets \Call{coreference\_resolution}{\mathcal{D}}$
        \ForEach {sentence $s$ \textbf{in} $\mathcal{D}$}
            \State $T_s \gets \Call{dependency\_parse}{s}$
            \State $T_s \gets \Call{identify\_node\_types}{T_s}$
            \State $T_s \gets \Call{prune\_tree}{T_s}$
            \State $T_s \gets \Call{merge\_nodes}{T_s}$
            \State $\mathcal{T} \gets \mathcal{T} \cup \{ T_s \}$
        \EndFor
    
        \State \Comment{Initialize graph}
        \State $\mathcal{G} = \{ \mathcal{V}, \mathcal{E} \}, \mathcal{V} \gets \emptyset, \mathcal{E} \gets \emptyset$ 
        \ForEach {tree $T = (V_T, E_T)$ \textbf{in} $\mathcal{T}$}
            \State $\mathcal{V} \gets \mathcal{V} \cup \{ V_T \}$
            \State $\mathcal{E} \gets \mathcal{E} \cup \{ E_T \}$
        \EndFor
        
        \State \Comment{Connect similar nodes}
        \ForEach {node $v_i$ \textbf{in} $\mathcal{V}$}
            \ForEach {node $v_j$ \textbf{in} $\mathcal{V}$}
                \If{$i \neq j$ \textbf{and} \Call{Is\_Similar}{$v_i$, $v_j$}}
                    \State $\mathcal{E} \gets \mathcal{E} \cup \{ \langle v_i, r^s, v_j \rangle, \langle v_j, r^s, v_i \rangle \}$
                \EndIf
            \EndFor
        \EndFor \\
        \Return {$\mathcal{G}$}
    \end{algorithmic}
\end{algorithm}

\subsection{DP-based Semantic Graph}

Dependency Parsing (DP) analyzes the grammatical structure of a sentence, establishing relationships between ``head" words and words that modify them, in a tree structure. Given a document $\mathcal{D}$ with $n$ sentences $\{s_1, \cdots, s_n\}$, Algorithm 2 gives the detailed procedure of constructing the semantic graph based on dependency parsing. 

To better represent the entity connection within the document, we first employ the coreference resolution system of AllenNLP to replace the pronouns that refer to the same entity with its original entity name. For each sentence $s$, we employ the AllenNLP implementation of the biaffine attention model~\cite{DBLP:conf/iclr/DozatM17} to obtain its dependency parse tree $T_s$. Afterwards, we perform the following operations to refine the tree: 

\vspace{2mm}
\noindent $\bullet$ $\text{IDENTIFY\_NODE\_TYPES}$: each node in the dependency parse tree is a word associated with a POS tag. To simplify the node type system, we manually categorize the POS types into three groups: \textit{verb}, \textit{noun}, and \textit{attribute}. Each node is then assigned to one group as its node type. 

\noindent $\bullet$ $\text{PRUNE\_TREE}$: we then prune each tree by removing unimportant continents (\textit{e.g.}, punctuation) based on pre-defined grammar rules.
Specifically, we do this recursively from top to bottom where for each node $v$, we visit each of its child node $c$. If $c$ needs to be pruned, we delete $c$ and directly link each child node of $c$ to $v$. 

\noindent $\bullet$ $\text{MERGE\_NODES}$: each node in the tree represents only one word, which may lead to a large and noisy semantic graph especially for long documents. To ensure that the semantic graph only retains important semantic relations, we merge consecutive nodes that form a complete semantic unit. To be specific, we apply a simple yet effective rule: merging a node $v$ with its child $c$ if they form a consecutive modifier, \textit{i.e.}, both the type of $v$ and $c$ are \textit{modifier}, and $v$ and $c$ is consecutive in the sentence. 

\vspace{2mm}

After obtaining the refined dependency parse tree $T_s$ for each sentence $s$, we add intra-tree edges to construct the semantic graph by connecting the nodes that are similar but from different parse trees. For each possible node pair $\langle v_i, v_j \rangle$, we add an edge between them with a special edge type \textit{SIMILAR} (denoted as $r^s$) if the two nodes are similar, \textit{i.e.}, satisfying the same condition as described in Section~\ref{appsec:SRL}. 

\subsection{Examples}
\label{appsec:examples}

Figure~\ref{fig:appenix1} shows a real example for the DP- and SRL-based semantic graph, respectively. In general, DP-based graph contains less words for each node compared with the SRL-based graph, allowing it to include more fine-grained semantic relations. For example, \textit{a leading member of the Native American self-determination movement} is treated as a single node in the SRL-based graph. While in the DP-based graph, it is represented as a semantic triple $\langle$ \textit{a leading member}, \textit{pobj}, \textit{the Native American self-determination movement} $\rangle$. As the node is more fine-grained in the DP-based graph, this makes the graph typically more sparse than the SRL-based graph, which may hinder the message passing during graph propagation. 

In experiments, we have compared the performance difference when using DP- and SRL-based graphs. We find that although both SRL- and DP-based semantic graph outperforms all baselines in terms of BLEU 1-4, DP-based graph performs slightly better than SRL-based graph ($+3.3\%$ in BLEU-4). 

\end{document}